%% file: main.tex
\documentclass[11pt]{article}
\input{imports}
\input{macros}

\newcommand{\email}[1]{\textsf{#1}}

\title{Local Regularizers Are Not Transductive Learners}

\author{
Sky Jafar \\ University High School \\ Irvine, CA \\ 
\email{jafar.sky@gmail.com} \and 
Julian Asilis \\ USC \\ Los Angeles, CA \\ \email{asilis@usc.edu} \and 
Shaddin Dughmi \\ USC  \\ Los Angeles, CA \\ \email{shaddin@usc.edu} 
}

\date{}

\begin{document}

\maketitle

\begin{abstract}
We partly resolve an open question raised by \citet[COLT 2024]{asilis2024regularization,asilis-open-problem}: whether the algorithmic template of \emph{local regularization} --- an intriguing generalization of explicit regularization, a.k.a.~structural risk minimization --- suffices to learn all learnable multiclass problems. Specifically, we provide a negative answer to this question in the transductive model of learning. We exhibit a multiclass classification problem which is learnable in both the transductive and PAC models, yet cannot be learned transductively by any local regularizer. The corresponding hypothesis class, and our proof, are based on principles from cryptographic secret sharing. We outline challenges in extending our negative result to the PAC model, leaving open the tantalizing possibility of a PAC/transductive separation with respect to local regularization.
\end{abstract}

\input{introduction}

\input{preliminaries}

\input{GBDLS}

\input{local_regularization}

\input{PAC_challenges}

\input{conclusion}

\vspace{-0.2 cm}
\subsection*{Acknowledgments}
\vspace{-0.2 cm}

\noindent 
The authors thank -- in alphabetical order -- Siddartha Devic, Vatsal Sharan, and Shang-Hua Teng for many useful conversations regarding the role of regularization in learning. 
Part of this work was done during the course of a research program administered by Pioneer academics; Sky Jafar and Shaddin Dughmi thank Pioneer and its staff for their support.
Julian Asilis was supported by the National Science Foundation Graduate Research Fellowship Program under Grant No.\ DGE-1842487, and completed this work in part while visiting the Simons Institute for the Theory of Computing.
Shaddin Dughmi was supported by NSF Grant CCF-2432219. 

\newpage 
\bibliographystyle{plainnat}
\bibliography{refs.bib}

\end{document}

%% file: imports.tex
\usepackage{fullpage}
\usepackage{appendix}
\usepackage[sort, authoryear, round]{natbib}
\usepackage{fancyhdr}
\usepackage[format=plain,
            labelfont={it,it},
            textfont=it]{caption}
\usepackage[stable, bottom, flushmargin]{footmisc}
\usepackage{graphicx}
\usepackage{xcolor}    
\usepackage{breakcites}
\usepackage{amsmath,amssymb,amsthm}
\usepackage{nicefrac}

\definecolor{lightblue}{RGB}{0,102,204}
\usepackage[pagebackref, colorlinks=true, linkcolor=lightblue, citecolor=lightblue, urlcolor=lightblue]{hyperref}

\usepackage{pgfplots}
\pgfplotsset{compat = newest}
\usepackage{wrapfig}

\usepackage{enumerate}
\usepackage{xparse, expl3}
\usepackage{indentfirst}
\usepackage{float}
\usepackage{tikz}
\usepackage{tikzscale}
\usetikzlibrary{arrows.meta,arrows}
\usetikzlibrary{decorations.pathreplacing}
\usetikzlibrary{positioning,chains,fit,shapes,calc,shapes.misc}

\usepackage[capitalize]{cleveref}

\newtheorem{theorem}{Theorem}[section]
\newtheorem{proposition}[theorem]{Proposition}
\newtheorem{lemma}[theorem]{Lemma}

\newtheorem{definition}[theorem]{Definition}

\newtheorem{remark}[theorem]{Remark}

\crefname{theorem}{Theorem}{Theorems}
\crefname{proposition}{Proposition}{Propositions}
\crefname{lemma}{Lemma}{Lemmas}
\crefname{corollary}{Corollary}{Corollaries}
\crefname{definition}{Definition}{Definitions}
\crefname{example}{Example}{Examples}
\crefname{remark}{Remark}{Remarks}
\crefname{algorithm}{Algorithm}{Algorithms}
\crefname{equation}{Equation}{Equations}
\crefname{section}{Section}{Sections}
\crefname{subsection}{Section}{Sections}
\crefname{conjecture}{Conjecture}{Conjectures}

\usepackage{microtype}
\usepackage{hyperref}
\usepackage{graphicx}

\usepackage{bbm,bm}
\usepackage{xfrac}
\usepackage{algorithm}
\usepackage{algpseudocode}
\usepackage{float}

\usepackage{color-edits}
\addauthor[Julian]{j}{purple}
\addauthor[Sky]{s}{cyan}
\addauthor[Shaddin]{sd}{green}

%% file: macros.tex
\newcommand{\nc}{\newcommand}
\newcommand{\rnc}{\renewcommand}

\newtheorem{open}{Open Problem}

\rnc{\epsilon}{\varepsilon}

\nc{\R}{\mathbb R}
\nc{\Q}{\mathbb Q}
\nc{\Z}{\mathbb Z}
\nc{\N}{\mathbb N}

\let\E\relax
\DeclareMathOperator*{\E}{\mathbb{E}}

\DeclareMathOperator*{\argmin}{arg\,min}

\nc{\CA}{\mathcal A}
\nc{\CC}{\mathcal C} 
\nc{\CD}{\mathcal D}
\nc{\CF}{\mathcal F}
\nc{\CG}{\mathcal G}
\nc{\CH}{\mathcal H}
\nc{\CI}{\mathcal I} 
\nc{\CS}{\mathcal S} 
\nc{\CX}{\mathcal X}
\nc{\CY}{\mathcal Y}

\nc{\VC}{\mathrm{VC}}
\nc{\DS}{\mathrm{DS}}
\nc{\Nat}{\mathrm{Nat}}
\nc{\Graph}{\mathrm{Graph}}

\nc{\Hotp}{\CH_{\mathrm{otp}}}

\nc{\Cantor}{\CH_\infty}

\nc{\defn}[1]{\textbf{#1}}
\nc{\im}{\mathrm{im}}
\rnc{\t}[1]{\text{#1}}
\nc{\ERM}{\mathrm{ERM}}
\nc{\Trans}{\mathrm{Trans}}
\nc{\h}{\widehat}
\nc{\Unif}{\mathrm{Unif}}
\nc{\xtest}{x_{\mathrm{test}}}
\nc{\xfool}{x_{\mathrm{fool}}}
\rnc{\arg}{\, \cdot \,}

%% file: introduction.tex
\section{Introduction}\label{Section:Introduction}

Understanding the power of various algorithmic templates for supervised learning is a core concern of both computational and statistical learning theory. This understanding arguably serves two purposes: a descriptive one by explaining the success of approaches employed in practice, and a prescriptive one which conveniently circumscribes the search space of promising algorithms for the practitioner. Most appealing are algorithmic approaches which are simple, natural, mirror what is seen in practice, and are powerful enough to learn in a rich variety of settings.

The most compelling  success story in this vein is that of \emph{empirical risk minimization (ERM)}. In practice, ERM and its approximations (such as gradient descent) are the work-horses of machine learning. In theory, ERM characterizes learnability for both binary classification~\citep{vapnik1982estimation,BEHW89} and agnostic real-valued regression~\citep{alon1997scale}.\footnote{That said, the sample complexity of ERM is slightly suboptimal for binary classification~\citep{hanneke2016optimal}, and more significantly suboptimal for regression~\citep{vavskevivcius2023suboptimality}.}

In second place is perhaps \emph{Structural Risk Minimization (SRM)}, a template for generalizations of ERM which trade off empirical risk with a user-specified measure $\psi(\arg)$ of model complexity, often referred to as a \emph{regularizer}. In trading off the fit of a model $h$ with its complexity $\psi(h)$, SRM protects against overfitting by encoding a preference for ``simple" models, à la Occam's razor. Approaches such as ridge regression, Lasso, and other instantiations of SRM have seen much success in applied machine learning. On the theoretical front, ridge regression learns successfully for a large class of convex and smooth problems, and SRM in the abstract is known to characterize non-uniform learnability~\citep{shalev2014understanding}.

These natural and instructive algorithmic characterizations have been largely limited to small or low-dimensional label spaces as in regression and binary classification. The next frontier in this regard is the family of multiclass classification problems, where neither ERM, SRM, nor any ``simple" algorithmic approach is known to learn whenever learning is possible. As evidence that any such characterization must employ more sophisticated algorithmic templates than ERM or SRM, \cite{DS14} show that no \emph{proper} learner can succeed in general; i.e., an optimal learner must sometimes ``stitch together'' different hypotheses to form its predictor. 
Improper algorithms which are optimal (or near-optimal) for multiclass classification have been described through orienting \emph{one-inclusion graphs} \citep{rubinstein2006shifting,DS14,aden2023optimal}, and as a combination of \emph{sample compression} and \emph{list learning} \citep{brukhim2022characterization}. While lending remarkable structural insight into multiclass learning, neither approach is ``simple'' in any meaningful sense, nor reminiscent of approaches employed in practice.

Our starting point for this paper, and perhaps the closest thing to the sort of algorithmic characterization we seek for multiclass problems, is the recent work of \cite{asilis2024regularization}. They demonstrate that every multiclass problem can be learned by an \emph{unsupervised local SRM (UL-SRM)} learner, a generalization of SRM on two fronts. First, the complexity of a hypothesis $h$ is described \emph{locally} per test point $x$ via a \emph{local regularizer} $\psi(h,x)$. Second, this local regularizer $\psi(\arg, \arg)$ is derived from the unlabeled data in what resembles an unsupervised pre-training stage.\footnote{We note that both local regularization and unsupervised pre-training have been successfully employed in the theory and practice of machine learning --- see \cite{wolf2008local,prost2021learning,vavskevivcius2023suboptimality,geprovable,azoury2001relative,vovk2001competitive}.} The former generalization, locality, is what makes such a learner improper, and therefore appears indispensible by the impossibility result of \cite{DS14}. Whether the second generalization, which allows learning the regularizer from unlabeled data, can be dispensed with is less clear. This was posed as an open problem by \cite{asilis-open-problem}, where they conjecture that dependence on the unlabeled data is indeed necessary; i.e., that \emph{local structural risk minimization} --- a.k.a.~\emph{local regularization} --- is insufficient for learning all learnable multiclass problems. 

The present paper resolves this conjecture in the affirmative for the transductive model of learning: we construct a hypothesis class which is learnable in both the transductive and PAC models, and yet cannot be transductively learned by a local regularizer. Our hypothesis class can be viewed as a cryptographic generalization of the \emph{first Cantor class} of \cite{DS14}, incorporating ideas from \emph{secret-sharing}. Each of our hypotheses divides a large domain arbitrarily in half, with each half receiving its own label. We ensure that each hypothesis is uniquely identified by its two labels, which by itself suffices to render the class learnable. However, the two halves ``share a secret,'' such that witnessing any one label reveals next to nothing about the hypothesis or the identity of the other label. While this does not obstruct learnability in general, we show that it does obstruct local regularizers: a transductive learner with vanishing error must exhibit a ``cycle'' in the typical test point's preferences over hypotheses, and therefore cannot be a local regularizer.

Although we were unable to extend our result to the PAC model, we suspect that this is a failure of our proof techniques rather than of our construction. We conjecture therefore that our same hypothesis class is in fact not learnable by any local regularizer in the PAC model. This would be in keeping with the tight relationship between the transductive and PAC models: learnability is equivalent in the two models, with sample-efficient reductions in both directions~--- see the discussion in \citep{trans_equiv_pac?,shaddin_position} and references therein. Whereas the reduction from PAC to transductive learning preserves the form of the learner, the converse reduction does not, obstructing black-box extensions of our result to the PAC model. Non-black-box approaches also appear challenging, and we outline the difficulties in \Cref{Section:PAC-difficulties}. Irrespective of the outcome for the PAC model, we argue that our result for the transductive model is interesting in its own right due to the tight relationship to the PAC model in most other respects, the promising hypothesis class we design and employ for our proof, and the intricacies involved in proving our main theorem (which uses a coupling argument at its center). Furthermore, the prospect of separating the transductive and PAC models with respect to learnability by local regularizers --- in the event that our conjecture is misguided --- is even more tantalizing.

%% file: preliminaries.tex
\section{Preliminaries}\label{Section:Preliminaries}

\subsection{Notation}

For a natural number $d \in \N$, we denote $[d] = \{1, \ldots, d\}$. For $Z$ a set, we use $Z^*$ to denote the set of all finite sequences in $Z$, i.e., $Z^* = \bigcup_{n \in \N} Z^n$. In particular, $\{0,1\}^*$ denotes the set of all finite binary strings. For $A, B \in \{0, 1\}^d$ binary strings of equal length, $A \oplus B$ refers to their entrywise XOR. We use $\sigma_0(A)$ and $\sigma_1(A)$ to refer to the entries at which $A$ takes the value of 0 or 1, respectively. That is, $\sigma_0(A) = \big\{ i \in [d] : A(i) = 0 \big\}, \sigma_1(A) = \big\{ i \in [d] : A(i) = 1 \big\}$.
We use $e(i)$ to denote the $i$th standard basis vector, whose length will always be clear from context (i.e., the vector with a 1 in its $i$th entry and zeroes elsewhere). 
For a statement $P$, $[P]$ denotes the Iverson bracket of $P$, i.e., $[P] = 1$ when $P$ is true and 0 otherwise. For a function $f: A \to B$, we let $\im(f) = \{f(a) : a \in A\}$ denote the image of $f$ in $B$.

\subsection{Learning Theory}

Unlabeled datapoints $x$ are drawn from a \defn{domain} $\CX$ and labeled by an element of the \defn{label set} $y \in \CY$. Pairs $(x, y) \in \CX \times \CY$ are referred as labeled datapoints or \emph{examples}. A \defn{training set} or training sequence is a tuple of labeled datapoints $S \in (\CX \times \CY)^n$. When clear from context, we will refer to labeled and unlabeled datapoints simply as \emph{datapoints}. A function $f \colon \CX \to \CY$ is a \defn{classifier} or \defn{predictor}. A collection of classifiers $\CH \subseteq \CY^\CX$ is referred to as a \defn{hypothesis class}; one of its elements $h \in \CH$ is a \defn{hypothesis}. Throughout the paper we employ the 0-1 loss function $\ell_{0-1} \colon \CY \times \CY \to \R_{\geq 0}$ defined by $\ell_{0-1}(y, y') = \big[ y \neq y' \big]$.
When $S = (x_i, y_i)_{i \in [n]}$ is a training set and $f \in \CY^\CX$ a classifier, the average loss incurred by $f$ on $S$ is referred to as its \defn{empirical risk} and denoted $L_S(f)$. That is, 
\[ L_S(f) = \frac{1}{n} \sum_{i \in [n]} \ell_{0-1} \big( f(x_i), y_i \big). \]

A \defn{learner} is a function from training sets to classifiers, e.g., $\CA : (\CX \times \CY)^* \to \CY^\CX$. We focus on learning in the realizable case, for which the purpose of a learner is to deduce the behavior of a \emph{ground truth} function $h^* \in \CH$ given its behavior on a finite training set. More precisely, we adopt the transductive model of learning, as employed by \citet{haussler1994predicting} and originally introduced by \citet{vapnik1974theory} and \citet{vapnik1982estimation}. 

\begin{definition}
The \defn{transductive model} of learning is that in which the following sequence of steps take place: 
\begin{enumerate}
    \item An adversary selects a collection of $n$ unlabeled datapoints $S  = (x_{i})_{i\in [n]}$ and a hypothesis $h^* \in \CH$. 
    \item The unlabeled datapoints $S$ are displayed to the learner. 
    \item An index $j \in [n]$ is selected at uniformly at random. The learner then receives the training set $(x_i, h^*(x_i))_{i \neq j}$. 
    \item The learner is prompted to predict the label of $x_j$, i.e., $h^*(x_j)$.
\end{enumerate}
\end{definition}

We refer to the choice of $S$ and $h^*$ parameterizing the above steps as a \emph{transductive instance}, and often collect this information into a pair, i.e., $(S, h^*)$. The \defn{transductive error} incurred by a learner on an instance $(S, h^*)$ is its average loss at the test point $x_j$, over the uniformly random choice of $j \in [n]$, i.e.,
\[ L^{\Trans}_{S, h^*}(\CA) = \frac 1n \sum_{i \in [n]} \big[ \CA(S_{-i}, h^*)(x_i) \neq h^*(x_i) \big], \]
where $\CA(S_{-i}, h^*)(x_i)$ denotes $\CA$'s output on the sample consisting of datapoints in $S_{-i} = S \setminus x_i$ labeled by $h^*$. 
Naturally, a class is transductively learnable if it can be learned to vanishingly small error on increasingly large datasets. 

\begin{definition}\label{Definition:trans-learning}
A hypothesis class $\CH \subseteq \CY^\CX$ is \defn{transductively learnable} if there exists a learner $\CA$ and function $m \colon (0, 1) \to \N$ such that for any $\epsilon \in (0, 1)$ and $h^* \in \CH$, if $S \in \CX^*$ has length $|S| \geq m(\epsilon)$, then 
\[ L_{S, h^*}^{\Trans}(\CA) \leq \epsilon. \]
\end{definition}

The pointwise minimal function $m$ such that there exists a learner $\CA$ satisfying \Cref{Definition:trans-learning} is referred to as the \emph{transductive sample complexity} of learning $\CH$, denoted $m_{\Trans, \CH}$, and learners which attain this sample complexity are said to be \emph{optimal}. (For our purposes, however, it will suffice to consider the property of learnability, without emphasizing sample complexities.) Notably, the transductive model of learning bears intimate connections with the Probably Approximately Correct (PAC) model of learning \citep{valiant1984theory}, which considers underlying probability distributions and requires learners to perform well on  randomly drawn test points when trained on i.i.d.\ training points. In particular, there is an equivalence between sample complexities in both models (up to a logarithmic factor), and techniques from transductive learning have recently been employed to establish the first characterizations of learnability for both multiclass classification and realizable regression \citep{brukhim2022characterization, attias2023optimal}. 
We defer a dedicated discussion of the PAC model to \Cref{Section:PAC-difficulties}.

In the landscape of learning, perhaps the most fundamental learners are given by ERM and SRM, which operate by selecting a hypothesis $h$ in the underlying class $\CH$ with lowest empirical risk (possibly balanced with an inductive bias over hypotheses in $\CH$, in the case of SRM). 

\begin{definition}\label{Definition:ERM}
A learner $\CA$ is an \defn{empirical risk minimization} (ERM) learner for a class $\CH$ if for all samples $S$, 
\[ \CA(S) \in \argmin_{h \in \CH} L_{S}(h). \] 
\end{definition}

\begin{definition}\label{Definition:SRM}
Let $\CH$ be a hypothesis class. A \defn{regularizer} for $\CH$ is a function $\psi: \CH \to \R_{\geq 0}$.
A learner $\CA$ is a \defn{structural risk minimization} (SRM) learner for a class $\CH$ if there exists a regularizer $\psi$ such that for all samples $S$, 
\[ \CA(S) \in \argmin_{h \in \CH} \Big( L_{S}(h) + \psi(h) \Big). \]
\end{definition}

\subsection{Local Regularization}

In realizable binary classification, ERM learners are known to succeed on all learnable classes, and furthermore to attain nearly-optimal sample complexity \citep{vapnik1974theory,BEHW89,ehrenfeucht1989general}. For multiclass learning over arbitrary label sets, however, these learning strategies are known to fail as a consequence of \citet[Theorem~1]{DS14}. In particular, \citet{DS14} establish that there exist multiclass problems which can only be learned by \defn{improper learners}: learners which may emit classifiers outside of the underlying hypothesis class $\CH$. (Note that ERM and SRM learners, in being phrased as arg min's over $\CH$, are necessarily \defn{proper}.) More recently, \citet{asilisunderstanding} expanded upon this result by demonstrating that there exist learnable multiclass problems which cannot be learned by any \emph{aggregation} of a finite number of proper learners, ruling out such strategies as majority voting of ERM learners, which has recently found success in binary classification \citep{aden2024majority,hogsgaardmany}. 

As such, multiclass learning in full generality requires different algorithmic blueprints than ERM and SRM. Perhaps the simplest such blueprint which has been considered is that of \emph{local regularization}, as introduced by \citet{asilis2024regularization}. Put simply, local regularization augments the regularizer --- usually a function $\psi \colon \CH \to \R_{\geq 0}$ --- to additionally receive an unlabeled datapoint as input, i.e., $\psi \colon \CH \times \CX \to \R_{\geq 0}$. The unlabeled datapoint is taken to be the test datapoint at which the learner is tasked with making a prediction. At each test datapoint $x \in \CX$, the learner then takes the behavior of the hypothesis $\widehat{h}$ with minimal value of $\psi(\widehat{h}, x)$, subject to $L_S(\widehat{h}) = 0$. 

This bears two crucial advantages over classical regularization.
\begin{itemize}
    \item[1.] Local regularization models the \emph{location-dependent} complexity of a hypotheses. It may be that $h$ acts as a simple function on a region $U \subseteq \CX$ yet as a complex function on $V \subseteq \CX$ (and vice versa for $h' \in \CH$). A local regularizer $\psi$ can model this behavior as $\psi(h, u) < \psi(h', u)$ for $u \in U$ and $\psi(h, v) > \psi(h', v)$ for $v \in V$. A classical regularizer, in contrast, is obligated to assign each of $h$ and $h'$ with a single value encoding their global complexity. 

    \item[2.] Local regularizers can induce improper learners, by emitting classifiers that ``stitch together" various hypotheses of $\CH$. In the previous case, for instance, one can imagine a learner $\CA$ induced by a local regularizer which takes the behavior of $h$ on $U \subseteq \CX$ and of $h'$ on $V \subseteq \CX$. 
\end{itemize}

\begin{definition}[{\cite{asilis-open-problem}}]\label{Definition:local-regularizer}
A \defn{local regularizer} for a hypothesis class $\CH$ is a function $\psi \colon \CH \times \CX \to \R_{\geq 0}$. A learner $\CA$ is said to be \defn{induced} by $\psi$ if for all samples $S$ and datapoints $x \in \CX$, 
\[ \CA(S)(x) \in \Big\{ h(x) : h \in \argmin_{h \in L_{S}^{-1}(0)} \psi(h, x) \Big\}. \]
We say that $\psi$ \defn{transductively learns} $\CH$ if all learners it induces are transductive learners for $\CH$. 
\end{definition}

\begin{remark}
Restricting to the set of hypotheses which incur zero empirical error in \Cref{Definition:local-regularizer}, rather than minimizing the sum of hypotheses' empirical error and regularization value, has the effect of simplifying the analysis of regularization. Furthermore, simply normalizing a regularizer to take outputs in the range $[0, \nicefrac{1}{n}]$ has the effect of equating the two perspectives for samples of size at most $n$. (Though this procedure is not uniform with respect to all possible sample sizes.) 
\end{remark}

\citet{asilis-open-problem} posed the open problem of whether local regularization is sufficiently expressive to learn all multiclass problems possible. 

\begin{open}[{\cite{asilis-open-problem}}]\label{Open-problem}
In multiclass classification, can all learnable hypothesis classes be learned by a local regularizer? If so, with optimal (or nearly optimal) sample complexity?
\end{open}

In Theorem~\ref{Theorem:local-regularization-fails} we resolve Open Problem~\ref{Open-problem} for the transductive model of learning, by exhibiting a learnable class which cannot be transductively learned by any local regularizer. In \Cref{Section:PAC-difficulties} we discuss the possibility of extending this result to the PAC model, and describe some of the challenges involved. 

\subsection{Secret Sharing}

We now provide a brief review of topics in \emph{secret sharing}, as the proof of our primary result, Theorem~\ref{Theorem:local-regularization-fails}, employs a hypothesis class whose structure is intimately related to concepts from the field.

\defn{Secret sharing} refers to the fundamental task of distributing private information, a \emph{secret}, among a group of \emph{players}. A successful solution consists of a technique for distributing information from a single \emph{dealer} to each player in such a manner that any individual player is incapable of recovering the secret, yet it can be revealed when the group works in concert. The information revealed to each player individually is referred to as a \emph{share}. In the general setting, there are $n$ players and the secret should be recoverable by any group of $t$ cooperating players, but not by any smaller group. (More precisely, any smaller group should not even be able to deduce partial information concerning the secret.) A solution to this problem is referred to as a $(t, n)$-threshold scheme.\footnote{In an even-more-general setting, the problem is defined by a collection of \emph{accessor subsets}, i.e., sets of players that should be able to deduce the secret when cooperating \citep{ito1989secret}.}

The study of secret sharing dates to the work of \citet{shamir1979share} and \citet{blakley1979safeguarding}; for a contemporary introduction, see \citet{beimel2011secret} or \citet{krenn2023introduction}. \citet{shamir1979share} designed an elegant $(t, n)$-threshold scheme which proceeds as follows: Let $q > n$ be a sufficiently large prime to contain all possible secrets, $k \in [q]$ the secret, and select $a_1, \ldots, a_{t-1}$ uniformly at random from $[q]$. Lastly, define the polynomial $P(x) = k + \sum_{i=1}^{t-1} a_i x^i \mod q$, and distribute to the $j$th player the share $(j, P(j))$. Then the cooperation of any $t$ players suffices to reveal $P$ (and thus $k$) owing to Lagrange's interpolation theorem, yet any smaller collection of players can reveal no information concerning $k = P(0)$ owing to the uniformly random choices of $a_1, \ldots, a_{t-1}$. 

For our purposes, however, it will suffice to consider the basic case of $t = n = 2$, for which there is a strikingly simple secret-sharing method referred to as the \defn{one-time pad} (OTP) in cryptography. Namely, given a secret $C \in \{0, 1\}^n$, the one-time pad selects a string $A$ uniformly at random from $\{0, 1\}^n$, and distributes to player one the share $A$ and to player two the share $A \oplus C$. Individually, then, each player witnesses a uniformly random string of length $n$, yet the XOR of their shares is precisely the secret, $A \oplus (A \oplus C) = C$. In \Cref{Section:local-regularization-fails}, we design a hypothesis class inspired by the one-time pad, for which --- roughly speaking --- each hypothesis $h$ is represented by a secret and each of its two possible outputs reveals one share of the secret. Crucially, then, the information of one of $h$'s secrets (i.e., its behavior on one half of the domain) maintains the identity of its remaining secret completely opaque (i.e., its behavior on the remaining half of the domain).\footnote{Strictly speaking, for the hypothesis class we construct, one of $h$'s ``secrets" and its behavior on a region of the domain can reveal partial information concerning its other ``secret."}

%% file: GBDLS.tex
\section{GBDLS Classes}\label{Section:GBDLS}

We devote this section to the development of two properties (of a hypothesis class) which will play a central role in the proof of \cref{Theorem:local-regularization-fails}: that of being \emph{generalized binary} and of having \emph{distinct label sets}. Upon defining each such property (of an underlying hypothesis class), we will demonstrate that neither is individually sufficient to ensure learnability. However, we show that all classes which have both properties --- termed \emph{generalized binary with distinct label sets} (GBDLS) --- are necessarily learnable. Neither observation is particularly difficult to prove, but serves towards the eventual proof of \cref{Theorem:local-regularization-fails}, and we believe that the language of such hypothesis classes may be of independent interest. 

\begin{definition}
A hypothesis $h \colon \CX \to \CY$ is \defn{k-ary} if $|\im(h)| \leq k$. When $k = 2$, we say that $h$ is a binary hypothesis. 
\end{definition}

\begin{definition}
A hypothesis class $\CH \subseteq \CY^\CX$ is \defn{k-ary} if $|\bigcup_{h \in \CH} \im(h)| \leq k$. When $k = 2$, we say that $\CH$ is binary. 
\end{definition}

We have slightly overloaded the term \emph{k-ary}, but there should be little risk of confusion: a hypothesis $h$ is $k$-ary if and only if the class $\{h\}$ is $k$-ary. 

\begin{definition}
A hypothesis class is \defn{generalized k-ary} if each $h \in \CH$ is a $k$-ary hypothesis. When $k = 2$, we say that $\CH$ is generalized binary.
\end{definition}

\begin{definition}
A hypothesis class $\CH$ \defn{has distinct label sets} if $\im \colon \CH \to 2^{\CY}$ is an injection. That is, $\im(h) \neq \im(h')$ when $h \neq h'$ and $h, h' \in \CH$. 
\end{definition}

It is not difficult to exhibit counter-examples demonstrating that neither the property of being generalized binary nor the property of having distinct label sets is sufficient to ensure that a hypothesis class $\CH$ is learnable. For the former, take any binary class of infinite VC dimension. For the latter, take any nonlearnable class $\CH \subseteq \CY^\CX$ and expand each of its hypotheses $h \in \CH$ to be defined on an additional unlabeled datapoint $*$ by $h(*) = h|_{\CX}$. 

However, we refer to classes with both properties as being \emph{generalized binary with distinct label sets} (GBDLS), and we now demonstrate that GBDLS classes are always learnable. 

\begin{proposition}\label{Proposition:GBDLS-means-learnable}
Let $\CH \subseteq \CY^\CX$ be a GBDLS hypothesis class. Then $\CH$ is learnable, in both the PAC and transductive models. 
\end{proposition}
\begin{proof}
We will demonstrate that $\CH$ cannot DS-shatter any sequence of 3 distinct points, from which it follows that $\DS(\CH) \leq 2$ and thus that $\CH$ is learnable in both models \citep{brukhim2022characterization}. Fix any sequence of distinct points $S = (x_1, x_2, x_3) \subseteq \CX$, and let $\CF \subseteq \CH|_S$ be a finite non-empty set of behaviors of $\CH$ on $S$. It remains to show that there exists an $f \in \CF$ and $i \in [3]$ for which $f$ does not have an $i$-neighbor in $\CF$ (i.e., a $g \in \CF$ with $g|_{S \setminus \{x_i\}} = f|_{S \setminus \{x_i\}}$ and $g(i) \neq f(i)$). 

To this end, select an arbitrary behavior $h \in \CF$, and identify it with the sequence $(h(x_j))_{j \in [3]} \subseteq \CY$. Suppose that $h$ is a constant behavior, i.e., that $h = (a, a, a)$ for a label $a \in \CY$. If $h$ does not have a 3-neighbor $f \in \CF$, then we are done. Otherwise, consider the 3-neighbor $f = (a, a, b)$ of $h \in \CF$. Note that if $h$ had been a non-constant behavior, then it would have taken the form $h = (a, a, b)$ to begin with (up to re-ordering of the points $(x_1, x_2, x_3)$, owing to the fact that $h$ is a binary behavior). We may thus consider a behavior $f = (a, a, b) \in \CF$ where $a \neq b$, without loss of generality. Then it follows immediately that $f$ does not have a 1-neighbor $g \in \CF$. Any such $g$ must take the form $g = (b, a, b)$, as $\CH$ is generalized binary, and this yields that $\im(f) = \{a, b\} = \im(g)$, producing contradiction with the fact that $\CH$ has distinct label sets. 
\end{proof}

%% file: local_regularization.tex
\section{Insufficiency of Local Regularization in the Transductive Model}\label{Section:local-regularization-fails}

We are now equipped to prove the primary result of the paper: There exists a learnable hypothesis class $\Hotp$ which cannot be transductively learned by any local regularizer (\cref{Theorem:local-regularization-fails}). We begin by defining the class $\Hotp$ which witnesses this separation. Notably, $\Hotp$ is intimately related to secret sharing techniques, including the \emph{one-time pad}. More precisely, each hypothesis $h \in \Hotp$ is parameterized by two strings $A, B \in \{0, 1\}^*$, and $h(i) = (0,A)$ or $h(i) = (1,B)$ depending upon the $i$th bit of $A \oplus B$. When a training set $S$ contains two distinct labels, learning is trivial: only the ground truth function attains zero training error. When the data distribution places full measure on a single label, however,  correctly predicting an unseen test point $j$ requires a local regularizer which favors functions mapping $j$ to that same label. (And, owing to the structure of the one-time pad, knowledge of $A \oplus B$ at the training points and complete knowledge of $A$ reveals little information regarding $A \oplus B$ at an unseen test point.) Using a coupling argument, we demonstrate that no local regularizer can do so simultaneously for all labels. 

Recall that for binary strings $A, B \in \{0, 1\}^n$, $A \oplus B$ denotes their entrywise XOR. We set $\sigma_0(A) = \{i \in [n] : A(i) = 0\}$ and likewise $\sigma_1(A) = \{i \in [n] : A(i) = 1\}$. In the event that $|\sigma_0(A)| = |\sigma_1(A)|$, $A$ is said to be \emph{balanced}. 

\begin{definition}\label{Definition:secret-sharing-class}
Let $\CX = \N$ and $\CY = \{0, 1\} \times \{0, 1\}^*$. For each $d \in \N$ and $A, B \in \{0, 1\}^d$, define 
\begin{align*}
h_{A, B} \colon \CX &\longrightarrow \CY \\
x &\longmapsto \begin{cases} (0, A) & (A \oplus B)(x \bmod d) = 0, \\ (1, B) & (A \oplus B)(x \bmod d) = 1. \end{cases} 
\end{align*} 
Then the hypothesis class $\Hotp \subseteq \CY^\CX$ is defined as
\[ \Hotp = \Big \{ h_{A, B} : A, B \in \{0, 1\}^*, |A| = |B|, A \oplus B \text{ is balanced} \Big \}.  \] 
\end{definition}

\begin{remark}
For those familiar with the \emph{first Cantor class} of \citet{DS14}, denoted $\Cantor$, note that $\Hotp$ can be seen as generalizing this class. In particular, $\Cantor$ is defined by ``glueing together" the classes $\{ \CH_d \}_{d \in \N}$, where $\CH_d \subseteq \CX_d^{\CY_d}$, $\CX_d$ is a set of size $d$, $\CY_d = 2^{\CX_d} \cup \{*\}$, and 
\[ \CH_d = \big \{ h_A : A \subseteq \CX_d, |A| = d / 2 \big \} \]
where 
$h_A(x) = A$ if $x \in A$ and $*$ otherwise. 
Each such $\CH_d$ can equivalently be defined in the language of Definition~\ref{Definition:secret-sharing-class} by identifying each $A \subseteq \CX_d$ with its characteristic vector, setting $h_A := h_{A, 1^d}$, and considering the relabeling $(1, 1^d) \mapsto *$, $(0, A) \mapsto A$. 
\end{remark}

\begin{lemma}
$\Hotp$ is a GBDLS class.
\end{lemma}
\begin{proof}
It is immediate from Definition~\ref{Definition:secret-sharing-class} that each $h \in \CH$ has $|\im(h)| = 2$, meaning $\CH$ is generalized binary. Furthermore, if $f \in \CH$ is such that $\im(f) = \{(0, A), (1, B)\}$ then it must be that $f = h_{A, B}$. Thus $\Hotp$ has distinct label sets. 
\end{proof}

We first equate the task of learning with a local regularizer to learning with a local regularizer which is \emph{locally injective}, i.e., injective on $\CH$ for each fixed choice of test point $x \in \CX$. 

\begin{definition}
A local regularizer $\psi \colon \CH \times \CX \to \R_{\geq 0}$ is \defn{locally injective} if $\psi(\arg, x)$ is injective for each $x \in \CX$. That is, $\psi(h, x) \neq \psi(h', x)$ for any $x \in \CX$ and $h \neq h' \in \CH$. 
\end{definition}

The following lemma establishes that for countable hypothesis classes $\CH$, their local regularizers $\psi$ can be assumed to be locally injective (i.e., to totally order $\CH$ at each location $x \in \CX$, rather than merely partially order). Note that in this case, $\psi$ induces a unique learner, as there is no ``tie-breaking" left to its induced learners. 

\begin{lemma}\label{Lemma:locally-injective-regularizer}
Let $\CH \subseteq \CY^\CX$ be a countable hypothesis class. Then $\CH$ can be learned by a local regularizer if and only if it can be learned by one which is locally injective. 
\end{lemma}
\begin{proof}
The backward direction is immediate. For the forward, let $\psi$ be a local regularizer which learns $\CH$, meaning that each of its induced learners succeeds on $\CH$. Recall that for each $x \in \CX$, $\psi( \cdot, x)$ defines a (strict) partial order over $\CH$, i.e., with $h < h'$ if $\psi(h, x) < \psi(h', x)$. For each such $x$, let $\overline{\psi(\cdot, x)}$ be any completion of this partial order into a total ordering. As $\CH$ is countable, $\overline{\psi(\cdot, x)}$ can be embedded into the real numbers. Define $\overline{\psi}(\cdot, x)$ using such an embedding, and consider the unique learner $\CA$ induced by $\overline{\psi}$. As $\overline{\psi}$ acts as a completion of $\psi$ at each $x \in \CX$, then $\CA$ is also a learner induced by $\psi$. Thus $\CA$ learners $\CH$, by our assumption on $\psi$, meaning $\overline{\psi}$ learns $\CH$. 
\end{proof}

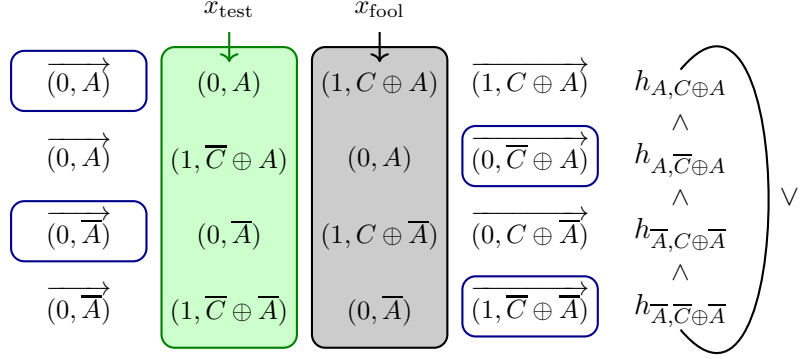
\begin{figure}[t] 
\centering
\input{figure.tikz}
\vspace{-40 pt}
\caption{Depiction of the learning problems $\{(S_i,h^*_i)\}_{i \in [4]}$ in \Cref{Theorem:local-regularization-fails}, with training sets $S_i$ circumscribed in blue. Each row corresponds to one of the learning problems $(S_i,h^*_i)$. Crucially, success of a local regularizer $\psi$ on each learning problem imposes the ordering relations on the right-hand side, which collectively produce a cycle.} \label{Figure:1}
\label{figure:bipartite}
\end{figure}

\begin{theorem}\label{Theorem:local-regularization-fails}
$\Hotp$ is a learnable hypothesis class, but cannot be transductively learned by any local regularizer.
\end{theorem}
\begin{proof}
Fix a local regularizer $\psi$ and $d \in \N$. By Lemma~\ref{Lemma:locally-injective-regularizer}, we may assume that $\psi$ is locally injective, inducing a unique learner $\CA$. Using a probabilistic argument, we will demonstrate that there exists a transductive learning instance $(S, h^*)$ with $|S| = 2d$ for which $\psi$ incurs error at least $\frac{1}{4}$. To this end, define the following independent random variables: 
\begin{itemize}
    \itemsep 0em
    \item $C$ is drawn uniformly at random from all balanced strings of length $2d$. 
    \item $A$ is drawn uniformly at random from $\{0, 1\}^{2d}$. 
    \item $m_0$ and $m_1$ are drawn uniformly at random from $[d]$. 
\end{itemize}
Further, define $\xtest$ to be the index of the $m_0$-th 0 entry in $C$, and $\xfool$ to be the index of the $m_1$-th 1 entry in $C$. (Note that such entries exist for any value of $m_0, m_1 \in [d]$, owing to the fact that $C$ is balanced.) 

We now define four random variables $\{T_i\}_{i \in [4]}$ based upon the previous variables. Each one is parameterized by a tuple $(S_i, h^*_i)_{i \in [4]}$ and measures the performance of $\CA$ at the test point $\xtest$ when trained the datapoints in $S_{i} \setminus \{\xtest\}$ labeled by $h^*_i$, the ground truth function. The $S_i$ and $h^*_i$ are defined as follows: 

\begin{itemize}
    \item[(1.)] Let $h^*_{1} = h_{A, C \oplus A}$, $S_1 = \sigma_0(C)$. Thus 
    \[  T_1 = \Big[\CA\big(S_1 \setminus \{\xtest\}, h^*_1\big)(\xtest) \neq h^*_1(\xtest) \Big],  \]
    where $\CA\big(S_1 \setminus \{\xtest\}, h^*_1\big)$ denotes the output of $\CA$ when trained on the dataset consisting of the points in $S_1 \setminus \{\xtest\}$ labeled by $h^*_1$. 
    
    \item[(2.)] Let $h^*_2 = h_{A, \overline{C} \oplus A}$, where $\overline{C}$ equals $C$ with its $\xtest$-th and $\xfool$-th entries each flipped.\footnote{Note that $\overline{C}$ is balanced, as $\xtest$ corresponds to a 0 entry in $C$ and $\xfool$ to a 1 entry.} Let $S_2 = \sigma_1(\overline{C}) = \sigma_1(C) \cup \{\xtest\} \setminus \{\xfool\}$. Thus 
    \[ T_2 = \big[\CA\big(S_2 \setminus \{\xtest\}, h^*_2\big)(\xtest) \neq h^*_2(\xtest) \big]. \]  
    
    \item[(3.)] Let $h^*_3 = h_{\overline{A}, C \oplus \overline{A}}$ where $\overline{A}$ denotes $A$ with its $\xtest$-th and $\xfool$-th entries each flipped. Let $S_3 = \sigma_0(C) = S_1$. Thus 
    \[ T_3 = \big[\CA\big(S_3 \setminus \{\xtest\}, h^*_3\big)(\xtest) \neq h^*_3(\xtest) \big]. \]  
    
    \item[(4.)] Let $h^*_4 = h_{\overline{A}, \overline{C} \oplus \overline{A}}$, $S_4 = \sigma_1(\overline{C}) = S_2$. Thus 
    \[ T_4 = \big[\CA\big(S_4 \setminus \{\xtest\}, h^*_4\big)(\xtest) \neq h^*_4(\xtest) \big]. \]  
\end{itemize}

\noindent It remains to prove the following lemma. 
\vspace{-0.7 cm}
\begin{quote}
\begin{lemma}\label{Lemma:internal-main-theorem}
Each random variable $T_i$, for $i \in [4]$, is distributed as the  error $\psi$ incurs on the (randomly-chosen) transductive learning instance $(S_i,h^*_i)$. Furthermore, for any value of the variables $C$, $A$, $m_0$, and $m_1$, the local regularizer $\psi$ must err at $\xtest$ in at least one of the instances $(S_i,h^*_i)_{i \in [4]}$. (That is, $\max_i T_i = 1$.)   
\end{lemma}
\begin{proof}
The first claim amounts to demonstrating that for each $i \in [4]$, conditioned upon $h^*_i$ and $S_i$, $\xtest$ is distributed uniformly randomly across $S_i$. For $i \in \{1, 3\}$, this is immediate. For $i \in \{2, 4\}$, recall that $S_2 = S_4 = \sigma_1(C) \cup \{\xtest\} \setminus \{\xfool\}$, and that $\xtest$ and $\xfool$ are independent and uniformly random elements of $\sigma_0(C)$ and $\sigma_1(C)$, respectively. Further, $C$ is chosen uniformly at random from the set of all balanced $2d$-strings. 
Then, conditioned upon $\overline{C}$, the probability that $\xtest$ takes the value of a given $s \in S = \sigma_1(\overline{C})$ is proportional to the cardinality of 
\begin{align*}
\Big\{ (\h{C}, \widehat{x}_{\mathrm{fool}}) &: s \in \sigma_0(\h{C}), \widehat{x}_{\mathrm{fool}} \in \sigma_1(\h{C}), \h{C} \text{ is balanced}, \h{C} \oplus e(s) \oplus e(\widehat{x}_{\mathrm{fool}}) = \overline{C} \Big\} \\
&= \Big\{ (\h{C}, \widehat{x}_{\mathrm{fool}}) : s \in \sigma_0(\h{C}), \widehat{x}_{\mathrm{fool}} \in \sigma_1(\h{C}), \h{C} \text{ is balanced}, \h{C} = \overline{C} \oplus e(s) \oplus e(\widehat{x}_{\mathrm{fool}}) \Big\}  \\
&= \Big\{ (\h{C}, \widehat{x}_{\mathrm{fool}}) : s \in \sigma_1(\overline{C}), \widehat{x}_{\mathrm{fool}} \in \sigma_0(\overline{C}), \h{C} \text{ is balanced}, \h{C} = \overline{C} \oplus e(s) \oplus e(\widehat{x}_{\mathrm{fool}}) \Big\}  \\
&= \Big\{ (\h{C}, \widehat{x}_{\mathrm{fool}}) : \widehat{x}_{\mathrm{fool}} \in \sigma_0(\overline{C}), \h{C}  = \overline{C} \oplus e(s) \oplus e(\widehat{x}_{\mathrm{fool}}) \Big\} \\
&\cong \Big\{ \widehat{x}_{\mathrm{fool}} : \widehat{x}_{\mathrm{fool}} \in \sigma_0(\overline{C}) \Big\}.
\end{align*}
The first equality follows from the observation that $\h{C} \oplus e(s) \oplus e(\widehat{x}_{\mathrm{fool}}) = \overline{C}$ is symmetric in $\h{C}$ and $\overline{C}$. The second equality rephrases the conditions on $s$ and $\widehat{x}_{\mathrm{fool}}$ in terms of $\overline{C}$, rather than $\h{C}$. The third equality employs the fact that $s \in \sigma_1(\overline{C})$ by definition, and that any $\h{C}$ satisfying the remaining conditions is automatically balanced. Thus the cardinality of this set does not depend upon $s$, completing the argument. 

For the second claim, note that $h_1^*(\xtest) \neq h_2^*(\xtest)$, as $C(\xtest) = 0$ yet $\overline{C}(\xtest) = 1$. However, $h_2^*$ attains zero empirical error on $S_1 \setminus \{\xtest\}$, as for any $s \in S_1  \setminus \{\xtest\}$,
\begin{align*}
h_1^*(s) = h_{A, C \oplus A}(s) = (0, A) = h_{A, \overline{C} \oplus A}(s) = h_2^*(s),
\end{align*}
where the first and third equalities use the fact that $s \in S_1 \setminus \{\xtest\} \subseteq \sigma_0(C) \cap \sigma_0(\overline{C})$. Thus, in order for $\psi$ to correctly classify task (1.), it must be that $\psi(h^*_1, \xtest) < \psi(h^*_2, \xtest)$. Likewise, $h_2^*(\xtest) \neq h_3^*(\xtest)$ because $C(\xtest) = 0 \neq 1 =  \overline{C}(\xtest)$, yet $h_3^*$ attains zero empirical error on $S_2 \setminus \{\xtest\}$, as $S_2 \setminus \{\xtest\} \subseteq \sigma_1(C) \cap \sigma_1(\overline{C})$ and $\overline{C} \oplus A = C \oplus \overline{A}$. Thus, for $\psi$ to correctly classify task (1.) would require that $\psi(h^*_2, \xtest) < \psi(h^*_3, \xtest)$. 

Invoke this reasoning twice more with $h^*_3$ and $h^*_4$; see \Cref{Figure:1}. In particular, $h^*_3(\xtest) \neq h^*_4(\xtest)$ as $C(\xtest) \neq \overline{C}(\xtest)$, yet $S_3 \setminus \{\xtest\} = \sigma_0(C) \setminus \{\xtest\} \subseteq \sigma_0(C) \cap \sigma_0(\overline{C})$. Thus $\psi$'s success on $T_3$ relies upon $\psi(h^*_3, \xtest) < \psi(h^*_4, \xtest)$. Finally, $h^*_4(\xtest) \neq h^*_1(\xtest)$, but $S_4 \setminus \{\xtest\} = \sigma_1(\overline{C}) \setminus \{\xtest\} \subseteq \sigma_1(C) \cap \sigma_1(\overline{C})$ and $C \oplus A = \overline{C} \oplus \overline{A}$. Thus success of $\psi$ on $T_4$ imposes the final requirement that $\psi(h^*_4, \xtest) < \psi(h^*_1, \xtest)$. It follows immediately that $\psi$ cannot succeed on all at once. 
\end{proof}
\end{quote}

By the second claim of Lemma~\ref{Lemma:internal-main-theorem}, the error $\psi$ incurs at $\xtest$, on average over the instances $(S_i, h^*_i)_{i \in [4]}$, is $\frac{1}{4}\sum_{i=1}^4 \E [T_i] \geq \frac{1}{4}$.  By the first claim of Lemma~\ref{Lemma:internal-main-theorem}
and a use of the probabilistic method, this implies the existence of a single transductive learning instance (on $2d$ points) for which $\psi$ incurs transductive error at least $\frac{1}{4}$. Conclude by recalling that $d$ was chosen arbitrarily. 
\end{proof}

%% file: figure.tikz
\definecolor{darkgreen}{rgb}{0.0, 0.5, 0.0}
\definecolor{lightgreen}{rgb}{0.8, 1.0, 0.8} 
\definecolor{gray}{rgb}{0.8, 0.8, 0.8} 
\definecolor{darkblue}{rgb}{0, 0, 0.545} 

\begin{tikzpicture}

    \def\xshiftA{-2.8}  
    \def\xshiftB{-0.8} 
    \def\xshiftC{1.2}  
    \def\xshiftD{3.2}  

    \draw[thick, darkgreen, rounded corners=5pt, fill=lightgreen] (\xshiftB-0.9,2.5) rectangle (\xshiftB+0.9,-1.5);

    \draw[thick, black, rounded corners=5pt, fill=gray] (\xshiftC-0.9,2.5) rectangle (\xshiftC+0.9,-1.5);

    \draw[thick, darkblue, rounded corners=5pt] (\xshiftA-0.9,2.45) rectangle (\xshiftA+0.9, 1.65);

    \draw[thick, darkblue, rounded corners=5pt] (\xshiftA-0.9,0.45) rectangle (\xshiftA+0.9, -0.35);

    \draw[thick, darkblue, rounded corners=5pt] (\xshiftD-0.9,2.45 - 1) rectangle (\xshiftD+0.9, 1.65 - 1);

    \draw[thick, darkblue, rounded corners=5pt] (\xshiftD-0.9,0.45 - 1) rectangle (\xshiftD+0.9, -0.35 - 1);

    \def\delta{0.07}
    \node at (\xshiftA, 2 + \delta) {\small $\overrightarrow{(0,A)}$};
    \node at (\xshiftA, 1 + \delta) {\small $\overrightarrow{(0,A)}$};
    \node at (\xshiftA, 0 + \delta) {\small $\overrightarrow{(0,\overline{A})}$};
    \node at (\xshiftA, -1 + \delta) {\small $\overrightarrow{(0,\overline{A})}$};
    
    \node at (\xshiftB,2) {\small $(0,A)$};
    \node at (\xshiftB,1) {\small $(1, \overline{C} \oplus A)$};
    \node at (\xshiftB,0) {\small $(0, \overline{A})$};
    \node at (\xshiftB,-1) {\small $(1, \overline{C} \oplus \overline{A})$};

    \node at (\xshiftC,2) {\small $(1, C \oplus  A )$};
    \node at (\xshiftC,1) {\small $(0, A)$};
    \node at (\xshiftC,0) {\small $(1, C \oplus \overline{A})$};
    \node at (\xshiftC,-1) {\small $(0, \overline{A})$};
    
    \node at (\xshiftD, 2 + \delta) {\small $\overrightarrow{(1,C \oplus A)}$};
    \node at (\xshiftD, 1 + \delta) {\small $\overrightarrow{(0,\overline{C} \oplus A)}$};
    \node at (\xshiftD, 0 + \delta) {\small $\overrightarrow{(0,C \oplus \overline{A})}$};
    \node at (\xshiftD, -1 + \delta) {\small $\overrightarrow{(1,\overline{C} \oplus \overline{A})}$};
        
    \node at (\xshiftB,3) {\small $\xtest$};
    \draw[->,darkgreen,thick] (\xshiftB,2.7) -- (\xshiftB,2.3);

    \node at (\xshiftC,3) {\small $\xfool$};
    \draw[->,black,thick] (\xshiftC,2.7) -- (\xshiftC,2.3);

    \pgfmathsetmacro{\xshiftE}{\xshiftD + 2}  

    \node at (\xshiftE, 2) { $h_{A, C \oplus A}$};
    \node at (\xshiftE, 1) { $h_{A, \overline{C} \oplus A}$};
    \node at (\xshiftE, 0) { $h_{\overline{A}, C \oplus \overline{A}}$};
    \node at (\xshiftE, -1) { $h_{\overline{A}, \overline{C} \oplus \overline{A}}$};
    
    \node at (\xshiftE, 1.5) {\small $\rotatebox{270}{$<$}$};
    \node at (\xshiftE, 0.5) {\small $\rotatebox{270}{$<$}$};
    \node at (\xshiftE, -0.5) {\small $\rotatebox{270}{$<$}$};
        
    \path[thick] (\xshiftE, -1.26) edge[out=-50, in=50, looseness=1.8] (\xshiftE, 2.2);
    \node at (\xshiftE + 1.45, 0.5) {$\lor$};

\end{tikzpicture}

%% file: PAC_challenges.tex
\section{Challenges in Extending to the PAC Model}\label{Section:PAC-difficulties}

Let us now describe by some of the challenges involved in extending \Cref{Theorem:local-regularization-fails} to the \emph{Probably Approximately Correct} (PAC) learning model of \citet{valiant1984theory}. We begin by recalling the model. 

\begin{definition}
Let $\CH \subseteq \CY^\CX$. A probability measure $\CD$ over $\CX \times \CY$ is \defn{$\CH$-realizable} if there exists an $h \in \CH$ for which
\[ L_\CD(h) := \E_{(x, y) \sim \CD} \big( \ell_{0-1}(h(x), y) \big) = 0. \] 
\end{definition}

\begin{definition}
Let $\CH \subseteq \CY^\CX$ be a hypothesis class and $\CA$ a learner. $\CA$ is said to be a \defn{PAC learner} for $\CH$ if there exists a function $m \colon (0, 1)^2 \to \N$ with the following property: for any $\CH$-realizable distribution $\CD$ and any $\epsilon, \delta \in (0, 1)$, if $S \sim \CD^{n}$ is a sample of size $n \geq m(\epsilon, \delta)$ drawn i.i.d.\ from $\CD$, then 
\[ L_\CD(\CA(S)) \leq \epsilon \] 
with probability at least $1 - \delta$ over the random choice of $S$. 
\end{definition}

In short, the PAC model differs from the transductive model by considering underlying probability distributions $\CD$ and requiring learners to attain favorable performance on $\CD$-i.i.d.\ test points when trained on $\CD$-i.i.d.\ training sets. Notably, the condition of learnability (by an arbitrary learner) is equivalent for both the PAC and transductive models. Furthermore, there are efficient reductions for converting optimal PAC learners into nearly-optimal transductive learners (and vice versa), demonstrating an equivalence between sample complexities in both models, up to only a logarithmic factor in the error parameter \citep{trans_equiv_pac?}. Converting a PAC learner $\CA$ into a transductive learner, however, requires randomly sampling from the training set $S$ and calling $\CA$ on the resulting dataset. This procedure does \emph{not} preserve the algorithmic form of transductive learners (as the resulting learner may even misclassify a point in $S$), and thus \Cref{Theorem:local-regularization-fails} does not imply --- at a black-box level --- the failure of local regularizers in the PAC model. 

We now describe three primary sources of difficulty in porting \Cref{Theorem:local-regularization-fails} to the PAC model. 

\begin{itemize}
    \item \textbf{Favoring non-ground-truth hypotheses.} The proof of \Cref{Theorem:local-regularization-fails}, and in particular of \Cref{Lemma:internal-main-theorem}, makes use of a train/test setting in which there are exactly two hypotheses attaining zero empirical error, only one of which --- the ground truth hypothesis --- is correct for the test point. Using this fact, we are able to deduce precise inequalities relating the various candidate ground truth hypotheses considered, which collectively lead to a contradiction. (I.e., to a cycle in the preferences of the local regularizer.) When learning in the PAC model, one immediately loses such tight control over the train/test setup, as both datasets are drawn i.i.d.\ from the underlying distribution $\CD$. In this setting, there will typically be various hypotheses attaining zero empirical error which are also correct at the test point, offering considerably more freedom to a successful local regularizer and (seemingly) prohibiting the detection of simple cycles. 

    \item \textbf{Incomparable version spaces.} Another approach may be  as follows: Note that a successful local regularizer $\psi$ for $\Hotp$ must, at a minimum, perform well on average when $A \in \{0, 1\}^n$ and $b \in \{0, 1\}$ are selected uniformly at random, the training set $S$ consist of $m < n$ points drawn uniformly at random from $[n]$ whose labels are all $(b, A)$, and the test point $\xtest$ is likewise a uniformly random point in $[n]$. (Whose correct label is $(b, A)$.) To deduce a contradiction from this strong condition imposed on $\psi$ requires considering its behavior on sets of ERM hypotheses $L_{S}^{-1}(0)$ for varying training sets $S$. (These sets are often referred to as the \emph{version spaces} of $S$ \citep{mitchell1977version}.) In almost all cases, distinct version spaces will be incomparable as sets (i.e., neither subsets nor supersets of one another), rendering it difficult to derive contradictions from such conditions. 

    \item \textbf{Error measurement.} One may note that by drawing $m$ points uniformly at random from a set $S = \big((x_i, y_i) \big)_{i \in [n]}$ for carefully chosen $m = \Theta (n \log n)$, it will occur with constant probability that exactly one point in $S$ is \emph{not} drawn. Na\"ively, this would seem to recover transductive learning as a special case of PAC learning. Crucially, however, the transductive model places full weight upon the learner's performance at the (unseen) test point, whereas the PAC model averages a learner's performance across the entire distribution $\CD = \Unif(S)$. In the PAC model, then, simply memorizing the training set suffices to learn $\Unif(S)$ to small error in $\Theta(n \log n)$ samples.  
\end{itemize}

%% file: conclusion.tex
\section{Conclusion}\label{Section:conclusion}

We study perhaps the simplest candidate template for multiclass learning: local regularization \citep{asilis-open-problem}. As our primary result, we demonstrate that there exists a learnable hypothesis class which cannot be transductively learned by any local regularizer. The hypothesis class $\Hotp$ which we employ for this result is based upon techniques from \emph{secret-sharing}, such as the one-time pad, and generalizes the \emph{first Cantor class} of \citet{DS14}. We conjecture that the same class also cannot be learned by any local regularizer in the PAC model, though we highlight some of the difficulties involved in extending our result in \Cref{Section:PAC-difficulties}.  